\documentclass[]{article}
\usepackage[letterpaper]{geometry}
\usepackage{amta2020}
\usepackage{times}
\usepackage{url}
\usepackage{latexsym}
\usepackage{natbib}
\usepackage{layout}

\usepackage{dirtytalk}
\usepackage{xcolor}
\usepackage{CJKutf8}

\begin{document}

% \amtaHeader{x}{x}{xxx-xxx}{2015}{45-character paper description goes here}{Author(s) initials and last name go here}
\title{\bf The Impact of Indirect Machine Translation on Sentiment Classification}

\author{
\name{\bf Alberto Poncelas} \hfill \addr{alberto.poncelas@adaptcentre.ie}\\
\name{\bf Pintu Lohar} \hfill\addr{pintu.lohar@adaptcentre.ie}\\
\name{\bf Andy Way} \hfill \addr{andy.way@adaptcentre.ie}\\
        \addr{ADAPT Centre, Dublin City University, Glasnevin, Dublin 9, Ireland}
\AND
       \name{\bf James Hadley} \hfill \addr{ hadleyj@tcd.ie}\\
        \addr{Trinity Centre for Literary and Cultural Translation and ADAPT Centre, Trinity College Dublin, Ireland}
}

\maketitle
\pagestyle{empty}

\begin{abstract}

Sentiment classification has been crucial for many natural language processing (NLP) applications, such as the analysis of movie reviews, tweets, or customer feedback. A sufficiently large amount of data is required to build a robust sentiment classification system. However, such resources are not always available for all domains or for all languages.

In this work, we propose employing a machine translation (MT) system to translate customer feedback into another language to investigate in which cases translated sentences can have a positive or negative impact on an automatic sentiment classifier. Furthermore, as performing a direct translation is not always possible, we explore the performance of automatic classifiers on sentences that have been translated using a pivot MT system.

We conduct several experiments using the above approaches to analyse the performance of our proposed sentiment classification system and discuss the advantages and drawbacks of classifying translated sentences.

\end{abstract}

\section{Introduction}
\label{sec:intro}

The key factor for building a sentiment classifier is training the model using a dataset of at least two different sentiment classes. This process requires a huge amount of data to build a workable sentiment classification system. However, sometimes it is difficult to find the required resources for the pertinent domain in large enough quantities and in all relevant languages. A typical and efficient approach to solve this problem consists in building the sentiment classification system with a high-resource language such as English and translating the sentences to be classified into this high-resource language.

An example of this is customer feedback classification. Many companies face difficulties analyzing what their clients say about their products and/or services because of the vast amount of feedback. This problem is aggravated for those businesses that have clients from several countries and therefore receive feedback in multiple languages. Although the feedback in multiple languages can be translated into English, the translated sentences may contain errors. Moreover, as the feedback consists of user-generated texts, it tends to be less grammatically strict than texts from literature, which implies that finding an accurate translation becomes more difficult ~\citep{lohar2019systematic}.

MT models are built to generate translations that carry the same meaning as the original sentence and are also fluent in the target language. However, in some scenarios, metrics for measuring fluency are less relevant. This is the case in the sentiment analysis of translated sentences, where maintaining the sentiment is the priority, even if the translation is not accurate in terms of \textit{adequacy} and \textit{fluency} ~\citep{tebbifakhr2019machine}. Despite the MT system generating understandable translations, it may not manifest the same sentiment as the original sentence. On top of that, in some cases, it is not possible to perform a direct translation, and so translation is required to be done via a pivot language. This may influence the classifier even more, as the errors produced by MT are propagated.

In this work, we analyze the difference in the performance of a sentiment classifier when using sentences in the original language, and sentences that have been translated (directly and indirectly) using an MT system. We discuss the benefits and disadvantages of using machine-translated sentences for automatic classification. 

The remainder of this paper is organised in the following manner. We discuss the related work done in this field in Section \ref{sec:related_work}. In Section \ref{sec:rq}, we formulate some research questions to be addressed in this area. The experiments are detailed in Section \ref{sec:exp}. We highlight our results in Section \ref{sec:res}. Finally, we conclude our present work in Section \ref{sec:conc}, followed by some possible future directions in Section \ref{sec:fut}. 

\section{Related Work}
\label{sec:related_work}

Several studies have addressed the issue of sentiment classification. The work in~\citet{pang-etal-2002-thumbs} examines the effectiveness of applying machine learning techniques to the sentiment classification of movie reviews. In ~\citet{li-etal-2010-sentiment} polarity shifting information is incorporated into a document-level sentiment classification system. First, polarity shifting is detected and then classifier combination methods are applied to perform polarity classification. However, in recent studies, deep learning-based approaches are gaining popularity for sentiment classification~\citep{zhang-etal-2019-aspect,Zhang2019aSentClass}.

MT plays a significant role in crosslingual sentiment analysis. An approach that is similar to ours is the work of~\citet{araujo2016evaluation}. Their experiments show that the performance of the English sentiment analysis tools on texts translated into English can be as good as using language-specific tools. Therefore, it may be worth deploying a system following the first approach, assuming some cost on the prediction performance. The work of ~\citet{barhoumi:hal-02042313} shows that the sentiment analysis of Arabic texts translated into English reaches a competitive performance with respect to standard sentiment analysis of Arabic texts. Using a high-quality MT system to translate a text from a specific language into English can eliminate the necessity of developing specific sentiment analysis resources for that language ~\citep{Shalunts2016TheIO}. One of the most recent approaches using MT for sentiment classification is described in ~\citet{tebbifakhr2019machine}. Their proposed approach for the sentiment classification of Twitter data in German and Italian shows that feeding an English classifier with machine-oriented translations improves its performance. For low-resource languages, MT-based approaches are considered efficient for analysing the sentiment of texts~\citep{kanayama-etal-2004-deeper,balahur-turchi-2012-multilingual}.

Additionally, several approaches aim to influence the MT to favour a sentiment when generating a translation.~\citet{lohar2017maintaining} propose training different SMT systems on sentences that have been tagged with a particular sentiment. Similarly, ~\citet{si2019sentiment} propose methods for generating translations of both positive and negative sentiments from the same sentence in the source language.

In our work, we not only investigate the sentiment classification on direct translation but also on indirect translation. Despite several existing studies on MT translation using a pivot language,
both in SMT ~\citep{utiyama2007comparison,wu2007pivot} and NMT~\citep{cheng2017joint,liu2018pivot}, to the best of our knowledge, this is the first study where indirect translation is explored for automatic sentiment classification.

\section{Research Questions}
\label{sec:rq}

In our experiments, we aim to explore the change in performance of a sentiment classifier when executed on MT-translated sentences. Furthermore, we want to compare the performance when using direct and indirect translation. The research questions (RQs) that we explore in this paper are the following:

\textbf{RQ1: To what extent do machine-translated sentences impact the performance of a sentiment classifier?}

Typically, MT models generate errors when producing translations. In addition, translating user-generated content (UGC) tends to be more difficult as it may contain spelling mistakes, wrong use of uppercase and lowercase letters, etc. Because of this, when customer feedback is translated the MT system may fail to produce a sentence with the same meaning or sentiment as the original sentence. We want to investigate how much MT errors affect the classification and whether they are proportional to the expected translation quality.

\textbf{RQ2: How much does the indirectly translated sentence impact a sentiment classifier?}

%In the situation where a direct translation is not possible (e.g. lack of resources for building an MT model or for building a classifier), the translation can be obtained indirectly using a pivot language.

In some situations, performing a direct translation into the language of the classifier is not possible. This is the case when language resources are available (e.g. either parallel data, training set for building a classification, etc.). Therefore the translation can be obtained indirectly using a pivot language. We can find many examples of languages that are generally translated via a pivot language. For instance, Irish is often translated via English, Basque and Catalan via Spanish, and Breton via French. The translation quality of a document that is indirectly translated is expected to be lower than a direct translation because the final translation accumulates the errors produced by two MT models.

This may also have a negative impact on the classifier. We want to analyze the performance of the classifier when classifying indirectly-translated sentences.

\section{Experiments}
\label{sec:exp}

%\paragraph{Data}
\subsection{MT settings}

We build an NMT system following the transformer approach ~\citep{vaswani2017attention} using OpenNMT~\citep{opennmt}. The model is trained for a maximum of $400$K steps using the recommended parameters,\footnote{\url{https://opennmt.net/OpenNMT-py/FAQ.html}} selecting the model that obtains the lowest perplexity on the development set.

A total of six translation models are built for translating French, Spanish and Japanese from/into English (two models for each pair).We use Paracrawl\footnote{\url{https://www.paracrawl.eu}} for English-French ($51$M parallel sentences) and English-Spanish ($39$M parallel sentences) language pairs, and JParaCrawl~\citep{morishita19jparacrawl} dataset ($8.7$M parallel sentences) for English-Japanese.

All the datsets are tokenized, truecased and then Byte Pair Encoding (BPE) ~\citep{sennrich2016neural} is applied with $89,500$ merge operations.

\begin{table}[!htbp]
\centering
\begin{center}
\begin{tabular}{ |p{2cm}|p{1.5cm}|}
\hline
	\textbf{Language}	&	\textbf{BLEU}		\\
\hline
En $\rightarrow$ Fr	&	31.77	\\
En $\rightarrow$ Es	&	40.13	\\
En $\rightarrow$ Ja	&	9.21	\\
\hline	
Fr $\rightarrow$ En	&	32.09	\\
Es $\rightarrow$ En	&	40.48	\\
Ja $\rightarrow$ En	&	12.85	\\
\hline
\end{tabular}
\caption{ 
MT performance
}
\label{table:BLEU_individual}
\end{center}
\end{table}

In order to estimate the performance of the MT models, in Table~\ref{table:BLEU_individual} we present the translation quality when translating a sample of $500$ lines from the \textit{news-commentary} dataset. The translations are evaluated using the BLEU~\citep{papineni2002bleu} metric. For English-French and English-Spanish language pairs the models achieve decent translation quality, but for the English-Japanese pair, BLEU scores are much lower. The reasons for this is that French and Spanish are grammatically and lexically closer to English than Japanese, and in addition, the number of training sentences used to build the MT system is four times smaller.

\subsection{Sentiment classifier}
\label{subsec:sentClass}

In order to build the sentiment analyser, we use the data from the IJCNLP-$2017$ \textit{Customer Feedback Analysis Task} ~\citep{liu2017ijcnlp}. This dataset consists of a collection of English, French, Spanish and Japanese  feedback containing short sentences extracted from reviews of products or services (in the hotel, restaurant or software domain). Each sentence is tagged with one or more categories based on the five-class system proposed by ~\citet{liu2017understanding}: comment, complaint, request, bug and meaningless. This is a fine-grained classification where only positive feedback is classified as comments whereas the other classes can be assumed to be different variants of negative feedback: \say{complaint} is defined as a negative comment and request or bug, which can also be considered negative as the client, or the user, is manifesting something that does not meet the standards of the product or service. In this dataset, the feedback can belong to multiple classes (e.g. a sentence such as \textit{my purchase won't show up.} would be classified as \say{complaint} and \say{bug}).

\begin{table}[!htbp]
\centering
\begin{center}
\begin{tabular}{ |p{0.75cm}|p{0.65cm}|p{0.57cm}|p{0.65cm}|p{0.57cm}|p{0.65cm}|p{0.57cm}|}
\hline
\textbf{Lang}	&	\multicolumn{2}{|c|}{\textbf{Train}  }	&	\multicolumn{2}{|c|}{\textbf{Dev}  }	&	\multicolumn{2}{|c|}{\textbf{Test}  }	\\
\hline
	&	Size	&	ratio	&	Size	&	ratio	&		Size	&	ratio		\\
\hline
En	&	3066	&	53\%	&	500	&	54\%	&	500	&	54\%	\\
Fr	&	1961	&	59\%	&	400	&	60\%	&	400	&	60\%	\\
Es	&	1632	&	61\%	&	300	&	81\%	&	300	&	76\%	\\
Ja	&	1527	&	52\%	&	250	&	55\%	&	300	&	56\%	\\
\hline					
\end{tabular}
\caption{Statistics of Feedback dataset. \textit{Size} indicates the number of lines, and \textit{ratio} the percentage of positive feedback.
}
\label{table:feedback_stats}
\end{center}
\end{table}

In order to simplify the analysis, we assume that feedback is positive if is classified as a \say{comment} only, and negative in all other cases. Table \ref{table:feedback_stats} describes the dataset. We see that the size of the training data ranges from $1.5K$ to $3K$ lines, depending on the language. In the \textit{ratio} column we indicate the percentage of feedback that is positive (and so $100-ratio$ would indicate the percentage of negative feedback). We can see that in all datasets positive feedback is predominant, but not enough to consider the dataset unbalanced.

We build the classifier using a bidirectional Gated Recurrent Unit (GRU) ~\citep{cho2014properties}, sigmoid activation function and a batch size of $16$. The classification is performed on data with BPE applied.

%\subsection{Baseline}

\begin{table}[!htbp]
\centering
\begin{center}
\begin{tabular}{ |p{1.5cm}|p{1.53cm}|p{1.5cm}|p{1.5cm}|}
\hline
	\textbf{Language}	&	\textbf{Accuracy}	&	\textbf{Precision}	&	\textbf{Recall}	\\
\hline
En	&	0.728    &	0.727	&	0.729	\\
Fr	&	0.788    &	0.779	&	0.783	\\
Es	&	0.856    &	0.798	&	0.852	\\
Ja	&	0.816    &	0.816	&	0.820	\\
\hline					
\end{tabular}
\caption{ 
Accuracy of the classifier on original sentences.
}
\label{table:baseline}
\end{center}
\end{table}

In Table \ref{table:baseline} we present the performance of the classifiers on the original sentences. We present three different metrics: accuracy, precision and recall. As we can see the three metrics have similar values. Therefore in the rest of this paper we will indicate only the accuracy (the number of correctly classified feedback over the total amount in the test set).

%\begin{itemize}
%    \item Accuracy: $(TP+TN)/(TP+TN+FP+FN)$
%    \item Precision: $(TP)/(TP+FP)$
%    \item Recall: $(TP)/(TP+FN)$
%\end{itemize}

%However, we can see that the three metrics
%, precision, and recall.

\section{Experiment Results}
\label{sec:res}

In our experiments we translate the test set into different languages and compare the performance of the classifier with the sentences classified in the original language (Table \ref{table:baseline}).

\subsection{Direct Translation Results}
\label{subsec:direct_translation}

We classify machine-translated feedback in the first set of experiments. These experiments can be divided into two parts: (i) Feedback translated into English; and (ii) Feedback translated from English.

\begin{table}[!htbp]
\centering
\begin{center}
\begin{tabular}{ |p{1.5cm}|p{1.5cm}||p{1.7cm}|}
\hline
	\textbf{Language}	&	\textbf{Accuracy}	&	\footnotesize \textbf{Accuracy (original language)}	\\
\hline
En $\rightarrow$ Fr	&	0.683	&	0.728	\\
En $\rightarrow$ Es	&	0.673	&	0.728	\\
En $\rightarrow$ Ja	&	0.573	&	0.728	\\
\hline	
Fr $\rightarrow$ En	&	0.754	&	0.788	\\
Es $\rightarrow$ En	&	0.805	&	0.856	\\
Ja $\rightarrow$ En	&	0.659	&	0.816	\\
\hline
\end{tabular}
\caption{Accuracy of the classifier on MT-generated sentences}
\label{table:individual}
\end{center}
\end{table}

In Table \ref{table:individual} we present the accuracy of the classifier when used on the translated data. The first column indicates in which direction the test set has been translated. The second column indicates the accuracy of the classifier when translated feedback is used. Therefore, we use the classifier built in the target language. Note that whereas in the first subtable the test set is the same (same English feedback translated into the other languages), in the second subtable each test set is different.

%\textcolor{red}{The classification we have been done one the translated feedback (target language) used the classifier trained in the language of the target (so we use the same language in which it has been trained). Note that whereas in the first subtable the test set is the same (same English feedbacks translated into the other languages), in the second subtable each test set is different.}

The accuracy of the classifier is lower for machine-translated data when compared to the sentences in the original sentences. In the first subtable the accuracies are lower than the original English classifier (\textit{En} row in Table \ref{table:baseline}). Similarly, in the second subtable, the accuracies are lower than those of \textit{Fr}, \textit{Es} and \textit{Ja} rows in Table \ref{table:baseline}.

We observe a relative decrease of between $4\%$ and $6\%$ of performance when using translations from/into French or Spanish. For example, 72.8\% of the English feedback is correctly classified by using the English classifier, but when these sentences are translated into French or Spanish and classified with the French and Spanish system, only 68.3\% and 67.3\% are correct. Similarly, in French and Spanish, 78.8\% and 85.6\% of the sentences are properly classified, but when translated into English this descend to 75.4\% and 80.5\%. The worst performance is seen in feedback translated from and into Japanese (around a $20\%$ of decrease). This is probably be related to the estimated translation quality shown in Table~\ref{table:BLEU_individual} where we see that MT models that use Japanese as source or target languages tend to perform worse. Nevertheless, the accuracy of the classifier and translation quality are not completely correlated. For example, when the English feedback is translated into French, classification accuracy decreases to $0.683$ (decrease of $6.2\%$) and when translated into Spanish accuracy decreases more (reaching an accuracy of $0.673$ which is a decrease of $7.6\%$) even though the translation quality should be better (according to Table \ref{table:BLEU_individual}).

In general, we observe that it is preferable to classify the sentences in the original language. Even when translating into a language with higher resources in which a classifier can be trained with more data (as is the case of translating Japanese feedback into English), the accuracy is lower. This may be related to translation quality. However we see that once a threshold of translation quality is achieved, it does not have a big impact on the classifier.

\subsection{Indirect Translation Results}
\label{subsec:indirect_translation}

In the second set of experiments we analyze the classification of indirectly-translated sentences, using English as a pivot language. We present in Table \ref{table:accuracy_indirect} the accuracy of the system with sentences translated indirectly. The rows indicate the source language and the columns the target language (after being translated from English).

\begin{table}[!htbp]
\centering
\begin{center}
\begin{tabular}{ |p{1cm}|p{1cm}|p{1cm}|p{1cm}|}
\hline
	\textbf{Lang.}	&	\textbf{Fr}	&	\textbf{Es}	&	\textbf{Ja}	\\
\hline
Fr	&	0.779    &	0.725    &	0.635	\\
Es	&	0.813    &	0.852    &	0.672	\\
Ja	&	0.667    &	0.697    &	0.726	\\
\hline					
\end{tabular}
\caption{Accuracy of classifier when using indirectly-translated sentences. The source language is that indicated in the row and the target language that indicated in the column. In all cases the pivot language is English.}
\label{table:accuracy_indirect}
\end{center}
\end{table}

We observe that the highest score in each row is the feedback that was translated back into the original language. In fact, the resulting indirect translation is similar to the original: $53.36$ BLEU points for French feedback; $56.63$ for Spanish; and $28.17$ for Japanese. \footnote{As BLEU evaluation metric is based on {\em n}-gram overlaps and Japanese sentences do not have whitespace separations between words, Japanese sentences were evaluated in BPE-split sentences} Although Japanese has a lower BLEU score, we observed that often the differences come from using a different writing system, e.g. both \begin{CJK}{UTF8}{min}おもう\end{CJK} (\textit{omou}) and \begin{CJK}{UTF8}{min}思う\end{CJK} (\textit{omou}) are the same word (\textit{to think}), but they are written in two different Japanese writing systems.

%in order to have {\em n}-grams instead of a single-unigram sentence

When comparing the results in Table \ref{table:accuracy_indirect} with those for direct translation (subtable at the bottom of Table \ref{table:individual}) we discover that the accuracy is similar. Furthermore, in some cases, indirectly-translated feedback becomes better classified. For example, when the Spanish feedback is translated into English, the accuracy of the classifier is $0.805$ whereas translating it further into French results in higher accuracy ($0.813$). Similarly, when the Japanese feedback is translated into English the accuracy is $0.659$ but when translated into French or Spanish the accuracy becomes $0.667$ and $0.697$, respectively.

As feedback consists of user-generated sentences they are expected to be informal, non-standard and noisy. Therefore, some of the terms may not be recognized by the classifier (which is trained using a small amount of data). However, in our experiments, the MT models are trained with larger amounts of data than the classifier and so they may recognize these terms and produce a less noisy translation.

In addition, we observe that many sentences in French, Spanish and Japanese use terms in English. For example, in Spanish feedback we find \textit{calles de shopping} (meaning \textit{shopping streets}) instead of \textit{calles comerciales}, or in Japanese we find \textit{staff\begin{CJK}{UTF8}{min} の接客\end{CJK}} (meaning \textit{staff reception}). The classifier in the original language may be unable to identify the words \textit{shopping} or \textit{staff} and this may affect the classification process. When performing translation, even if these OOV terms are copied directly into the target sentence, they will form a well-written sentence. When the feedback is further translated from the English pivot-sentences, the MT model is capable of translating the word.

\subsection{Translation Analysis}

In Table \ref{table:translation_examples} we present some translation examples that help us illustrate the problems and benefits of using MT-generated sentences in the classification. 

\begin{table*}[!htbp]
\centering
\begin{center}
\begin{tabular}{ |p{2cm}|p{5cm}||p{5cm}|}
\hline
	\textbf{Language}	&	\textbf{Sentence}	&	\textbf{Sentence (human-translated into English)}	\\
\hline
\small Ja	&	\begin{CJK}{UTF8}{min}ロケーション\end{CJK}	&	location	\\
\small Ja $\rightarrow$ En	&	location Location Location Location	&	location location location ...	\\
\small Ja $\rightarrow$ En $\rightarrow$ Ja	&	\begin{CJK}{UTF8}{min}場所 場所 場所 場所 場所 場所 場所\end{CJK}	&	location location location ...	\\
\hline
\small Es	&	seguro volveré	&	I will come back for sure	\\
\small Es $\rightarrow$ En	&	I will return insurance	&	I will return insurance	\\
\small Es $\rightarrow$ En $\rightarrow$ Es	&	voy a devolver el seguro	&	I will return insurance	\\
\hline
\small Fr	&	et le quartier pas très sympa.	&	and the neighborhood is not very nice.	\\
\small Fr $\rightarrow$ En	&	and the neighborhood is not very nice.	&	and the neighborhood is not very nice.	\\
\small Fr $\rightarrow$ En $\rightarrow$ Fr	&	et le quartier n'est pas très agréable.	&	and the neighborhood is not very nice.	\\
\hline
\end{tabular}
\caption{ 
Translation examples
}
\label{table:translation_examples}
\end{center}
\end{table*}

In the first subtable, we show how the MT engine produces a word-repetition (i.e. \textit{location}) when generating a translation. As the sentence is nonsense, it can be misinterpreted by the classifier. On top of that, the word-repetition problem is further replicated on indirect translation.

%This can be misinterpreted by the classifier as nonsense.

One of the problems is that indirect translation may produce errors that are propagated to the following MT system. In Table \ref{table:translation_examples} the Spanish sentence \textit{seguro volveré} is wrongly translated as \textit{I will return insurance} (the word \textit{seguro} can mean either \textit{for sure} or \textit{insurance}). This causes a positive sentence to become negative because of this error, and it is spread to the following translations.

In the second subtable, we show why in some cases using a translation could be beneficial. The original sentence in French \textit{et le quartier pas très sympa.} is not a grammatically-correct sentence as the word \textit{ne} (the negation of a verb in French follows the structure \say{ne}+VB+\say{pas}) and the verb \textit{est} (\textit{to be}) are omitted which is common in spoken French. The translation into English is accurate in meaning and when the sentence has been translated back to French, the structure of the sentence is correct. Another advantage is that MT-generated texts tend to have a lower lexical diversity ~\citep{toral2019post,vanmassenhove2019lost} which makes the classification easier. This can be seen with the French word \textit{sympa} which is not as frequent as \textit{agréable}. For example, there are $6,065$ occurrences of the word \textit{sympa} in the Paracrawl dataset whereas \textit{agréable} occurs $78,689$ times.

\section{Conclusions}
\label{sec:conc}

In this work we investigated the impact of both direct and indirect translation when evaluated in terms of sentiment preservation (rather than other common criteria such as \textit{adequacy} and \textit{fluency}). We performed translation of customer feedback and categorized it as positive or negative using an automatic classifier. There are several conclusions that we can draw from the experiments carried out.

\paragraph{Sentiment classifiers do not classify translated data as well as original sentences.} As expected, the outcome of our experiments shows that it is preferable to use the original feedback rather than a translation for classification. The MT-generated feedback introduces errors ~\citep{lohar2019systematic,nunez2019comparison} that causes the classifier to show worse performance.

\paragraph{Translation quality is not completely correlated to the performance of the classifier.}
Although the automatic sentiment classifier does not perform well on sentences with low-quality translation, after a certain translation-quality threshold the performance of the classifier is not correlated with the translation quality.

\paragraph{There are potential benefits to using MT-translated sentences}
Although MT models produce errors, they also tend to generate sentences with a lower amount of lexical translation, which facilitates the classification. Moreover, if the feedback, which is UGC, contains terms in the target language it may be easier for the classifier to classify the translated version.

\paragraph{The performance of the classifier on indirectly-translated sentences is similar to that when classifying directly-translated sentences.}
Despite the decrease in performance when classifying directly-translated feedback, we observe that it is similar to indirectly-translated feedback if the performance of the MT models is good enough. The sentences generated by the first MT system are expected to be of lower lexical diversity as compared to the user-generated sentences. This causes the second MT system in the pipeline to only have to translate simpler sentences.

\section{Future Work}
\label{sec:fut}

One of the limitations of our experiments is that we used an MT model to translate only the test set. An alternative experiment would involve translating the training data instead. The use of synthetic data for building models has been extensively explored in MT-field. Techniques such as back-translation ~\citep{sennrich2015improving,poncelas2018investigating}, in which synthetic data is created by translating sentences from another language, has proven to be useful for improving MT models. We want to explore whether using machine-generated sentences as training data for the classifier also has an impact on the performance.

In the future, we want to explore other experimental configurations. For example, in this paper we explored a classifier trained with a single language. We want to investigate whether the performance would be similar when using multilingual classifiers ~\citep{plank2017all}.

Moreover, in these experiments the MT models were trained on large amounts of data (9M to 51M sentences). Although smaller models are expected to produce lower quality translations, these may be enough for the sentiment classifier to achieve acceptable results. A future extension to this work would involve investigating what is the minimum amount of data necessary for building the MT system to create translations that are good enough for the classifier to perform well. Alternatively, small MT models can be built by selecting a subset of the available data \citep{silva2018extracting} that is closer to the user-generated content.

Another configuration would involve adapting the MT models to different categories. Following the approach of ~\citep{lohar2017maintaining} we could build different MT models for different classes. Alternatively, models could be adapted to translate feedback of a particular sentiment in a similar way to domain-adaptation. This can be done by fine-tuning with in-domain sets~\citep{van2017dynamic,poncelas2018feature} or appending a tag with the domain ~\citep{sennrich2016controlling,poncelas2019adapting}.

\section{Acknowledgements}

This research has been supported by the ADAPT Centre for Digital Content Technology which is funded under the SFI Research Centres Programme (Grant 13/RC/2106).

The QuantiQual Project, generously funded by the Irish Research Council’s COALESCE scheme (COALESCE/2019/117).

\small

\bibliographystyle{apalike}
\bibliography{amta2020}

\begin{thebibliography}{}

\bibitem[Araujo et~al., 2016]{araujo2016evaluation}
Araujo, M., Reis, J., Pereira, A., and Benevenuto, F. (2016).
\newblock An evaluation of machine translation for multilingual sentence-level
  sentiment analysis.
\newblock In {\em Proceedings of the 31st Annual ACM Symposium on Applied
  Computing}, pages 1140--1145, Pisa, Italy.

\bibitem[Balahur and Turchi, 2012]{balahur-turchi-2012-multilingual}
Balahur, A. and Turchi, M. (2012).
\newblock Multilingual sentiment analysis using machine translation?
\newblock In {\em Proceedings of the 3rd Workshop in Computational Approaches
  to Subjectivity and Sentiment Analysis}, pages 52--60, Jeju, Korea.

\bibitem[Barhoumi et~al., 2018]{barhoumi:hal-02042313}
Barhoumi, A., Aloulou, C., Camelin, N., Est{\`e}ve, Y., and Belguith, L.
  (2018).
\newblock {Arabic Sentiment analysis: an empirical study of machine
  translation's impact}.
\newblock In {\em {Language Processing and Knowledge Management International
  Conference (LPKM-2018)}}, Sfax, Tunisia.

\bibitem[Cheng et~al., 2017]{cheng2017joint}
Cheng, Y., Yang, Q., Liu, Y., Sun, M., and Xu, W. (2017).
\newblock Joint training for pivot-based neural machine translation.
\newblock In {\em Proceedings of the Twenty-Sixth International Joint
  Conference on Artificial Intelligence}, pages 3974--3980, Melbourne,
  Australia.

\bibitem[Cho et~al., 2014]{cho2014properties}
Cho, K., van Merri{\"e}nboer, B., Bahdanau, D., and Bengio, Y. (2014).
\newblock On the properties of neural machine translation: Encoder--decoder
  approaches.
\newblock In {\em Proceedings of SSST-8, Eighth Workshop on Syntax, Semantics
  and Structure in Statistical Translation}, pages 103--111, Doha, Qatar.

\bibitem[Kanayama et~al., 2004]{kanayama-etal-2004-deeper}
Kanayama, H., Nasukawa, T., and Watanabe, H. (2004).
\newblock Deeper sentiment analysis using machine translation technology.
\newblock In {\em {COLING} 2004: Proceedings of the 20th International
  Conference on Computational Linguistics}, pages 494--500, Geneva,
  Switzerland.

\bibitem[Klein et~al., 2017]{opennmt}
Klein, G., Kim, Y., Deng, Y., Senellart, J., and Rush, A.~M. (2017).
\newblock Opennmt: Open-source toolkit for neural machine translation.
\newblock In {\em Proceedings of the 55th Annual Meeting of the Association for
  Computational Linguistics-System Demonstrations}, pages 67--72, Vancouver,
  Canada.

\bibitem[Li et~al., 2010]{li-etal-2010-sentiment}
Li, S., Lee, S. Y.~M., Chen, Y., Huang, C.-R., and Zhou, G. (2010).
\newblock Sentiment classification and polarity shifting.
\newblock In {\em Proceedings of the 23rd International Conference on
  Computational Linguistics (Coling 2010)}, pages 635--643, Beijing, China.

\bibitem[Liu et~al., 2017a]{liu2017understanding}
Liu, C.-H., Groves, D., Hayakawa, A., Poncelas, A., and Liu, Q. (2017a).
\newblock Understanding meanings in multilingual customer feedback.
\newblock In {\em Proceedings of First Workshop on Social Media and User
  Generated Content Machine Translation ({{Social MT}} 2017)}, Prague, Czech
  Republic.

\bibitem[Liu et~al., 2017b]{liu2017ijcnlp}
Liu, C.-H., Moriya, Y., Poncelas, A., and Groves, D. (2017b).
\newblock {IJCNLP-2017 Task 4: Customer Feedback Analysis}.
\newblock In {\em Proceedings of the {{IJCNLP}} 2017, Shared Tasks}, pages
  26--33, Taipei, Taiwan.

\bibitem[Liu et~al., 2018]{liu2018pivot}
Liu, C.-H., Silva, C.~C., Wang, L., and Way, A. (2018).
\newblock Pivot machine translation using chinese as pivot language.
\newblock In {\em China Workshop on Machine Translation}, pages 74--85,
  Wuyishan, China.

\bibitem[Lohar et~al., 2017]{lohar2017maintaining}
Lohar, P., Afli, H., and Way, A. (2017).
\newblock Maintaining sentiment polarity in translation of user-generated
  content.
\newblock {\em The Prague Bulletin of Mathematical Linguistics}, 108(1):73--84.

\bibitem[Lohar et~al., 2019]{lohar2019systematic}
Lohar, P., Popovi{\'c}, M., Afli, H., and Way, A. (2019).
\newblock A systematic comparison between {{SMT}} and {{NMT}} on translating
  user-generated content.
\newblock In {\em 20th International Conference on Computational Linguistics
  and Intelligent Text Processing}, La Rochelle, France.

\bibitem[Morishita et~al., 2019]{morishita19jparacrawl}
Morishita, M., Suzuki, J., and Nagata, M. (2019).
\newblock {JParaCrawl}: A large scale web-based {{J}}apanese-{{E}}nglish
  parallel corpus.
\newblock {\em arXiv preprint arXiv:1911.10668}.

\bibitem[Nunez et~al., 2019]{nunez2019comparison}
Nunez, J. C.~R., Seddah, D., and Wisniewski, G. (2019).
\newblock Comparison between {{NMT}} and {{PBSMT}} performance for translating
  noisy user-generated content.
\newblock In {\em Proceedings of the 22nd Nordic Conference on Computional
  Linguistics (NoDaLiDa)}, pages 2--14, Turku, Finland.

\bibitem[Pang et~al., 2002]{pang-etal-2002-thumbs}
Pang, B., Lee, L., and Vaithyanathan, S. (2002).
\newblock Thumbs up? sentiment classification using machine learning
  techniques.
\newblock In {\em Proceedings of the 2002 Conference on Empirical Methods in
  Natural Language Processing ({EMNLP} 2002)}, pages 79--86, Philadelphia, USA.

\bibitem[Papineni et~al., 2002]{papineni2002bleu}
Papineni, K., Roukos, S., Ward, T., and Zhu, W.-J. (2002).
\newblock Bleu: a method for automatic evaluation of machine translation.
\newblock In {\em Proceedings of 40th Annual Meeting of the Association for
  Computational Linguistics}, pages 311--318, Philadelphia, Pennsylvania, USA.

\bibitem[Plank, 2017]{plank2017all}
Plank, B. (2017).
\newblock All-in-1 at ijcnlp-2017 task 4: Short text classification with one
  model for all languages.
\newblock In {\em Proceedings of the IJCNLP 2017, Shared Tasks}, pages
  143--148, Taipei, Taiwan.

\bibitem[Poncelas et~al., 2018a]{poncelas2018feature}
Poncelas, A., Maillette~de Buy~Wenniger, G., and Way, A. (2018a).
\newblock Feature decay algorithms for neural machine translation.
\newblock In {\em Proceedings of the 21st Annual Conference of the European
  Association for Machine Translation}, pages 239--248, Alicante, Spain.

\bibitem[Poncelas et~al., 2019]{poncelas2019adapting}
Poncelas, A., Sarasola, K., Dowling, M., Way, A., Labaka, G., and Alegria, I.
  (2019).
\newblock {Adapting NMT to caption translation in Wikimedia Commons for
  low-resource languages}.
\newblock {\em Procesamiento del Lenguaje Natural}, 63:33--40.

\bibitem[Poncelas et~al., 2018b]{poncelas2018investigating}
Poncelas, A., Shterionov, D., Way, A., de~Buy~Wenniger, G.~M., and Passban, P.
  (2018b).
\newblock Investigating backtranslation in neural machine translation.
\newblock In {\em 21st Annual Conference of the European Association for
  Machine Translation}, pages 249--258, Alicante, Spain.

\bibitem[Sennrich et~al., 2016a]{sennrich2016controlling}
Sennrich, R., Haddow, B., and Birch, A. (2016a).
\newblock Controlling politeness in neural machine translation via side
  constraints.
\newblock In {\em Proceedings of the 2016 Conference of the North American
  Chapter of the Association for Computational Linguistics: Human Language
  Technologies}, pages 35--40, San Diego, USA.

\bibitem[Sennrich et~al., 2016b]{sennrich2015improving}
Sennrich, R., Haddow, B., and Birch, A. (2016b).
\newblock Improving neural machine translation models with monolingual data.
\newblock In {\em Proceedings of the 54th Annual Meeting of the Association for
  Computational Linguistics (Volume 1: Long Papers)}, pages 86--96, Berlin,
  Germany.

\bibitem[Sennrich et~al., 2016c]{sennrich2016neural}
Sennrich, R., Haddow, B., and Birch, A. (2016c).
\newblock Neural machine translation of rare words with subword units.
\newblock In {\em Proceedings of the 54th Annual Meeting of the Association for
  Computational Linguistics (Volume 1: Long Papers)}, volume~1, pages
  1715--1725, Berlin, Germany.

\bibitem[Shalunts et~al., 2016]{Shalunts2016TheIO}
Shalunts, G., Backfried, G., and Commeignes, N. (2016).
\newblock The impact of machine translation on sentiment analysis.
\newblock In {\em The Fifth International Conference on Data Analytics},
  volume~63, pages 51--56, Venice, Italy.

\bibitem[Si et~al., 2019]{si2019sentiment}
Si, C., Wu, K., Aw, A., and Kan, M.-Y. (2019).
\newblock Sentiment aware neural machine translation.
\newblock In {\em Proceedings of the 6th Workshop on Asian Translation}, pages
  200--206, Hong Kong, China.

\bibitem[Silva et~al., 2018]{silva2018extracting}
Silva, C.~C., Liu, C.-H., Poncelas, A., and Way, A. (2018).
\newblock Extracting in-domain training corpora for neural machine translation
  using data selection methods.
\newblock In {\em Proceedings of the Third Conference on Machine Translation:
  Research Papers}, pages 224--231, Brussels, Belgium.

\bibitem[Tebbifakhr et~al., 2019]{tebbifakhr2019machine}
Tebbifakhr, A., Bentivogli, L., Negri, M., and Turchi, M. (2019).
\newblock Machine translation for machines: the sentiment classification use
  case.
\newblock In {\em 2019 Conference on Empirical Methods in Natural Language
  Processing and the 9th International Joint Conference on Natural Language
  Processing}, pages 1368--1374, Hong Kong, China.

\bibitem[Toral, 2019]{toral2019post}
Toral, A. (2019).
\newblock Post-editese: an exacerbated translationese.
\newblock In {\em Proceedings of Machine Translation Summit XVII Volume 1:
  Research Track}, pages 273--281, Dublin, Ireland.

\bibitem[Utiyama and Isahara, 2007]{utiyama2007comparison}
Utiyama, M. and Isahara, H. (2007).
\newblock A comparison of pivot methods for phrase-based statistical machine
  translation.
\newblock In {\em Human Language Technologies 2007: The Conference of the North
  American Chapter of the Association for Computational Linguistics;
  Proceedings of the Main Conference}, pages 484--491, Rochester, USA.

\bibitem[van~der Wees et~al., 2017]{van2017dynamic}
van~der Wees, M., Bisazza, A., and Monz, C. (2017).
\newblock Dynamic data selection for neural machine translation.
\newblock In {\em Proceedings of the 2017 Conference on Empirical Methods in
  Natural Language Processing}, pages 1400--1410, Copenhagen, Denmark.

\bibitem[Vanmassenhove et~al., 2019]{vanmassenhove2019lost}
Vanmassenhove, E., Shterionov, D., and Way, A. (2019).
\newblock Lost in translation: Loss and decay of linguistic richness in machine
  translation.
\newblock In {\em Proceedings of Machine Translation Summit XVII Volume 1:
  Research Track}, pages 222--232, Dublin, Ireland.

\bibitem[Vaswani et~al., 2017]{vaswani2017attention}
Vaswani, A., Shazeer, N., Parmar, N., Uszkoreit, J., Jones, L., Gomez, A.~N.,
  Kaiser, {\L}., and Polosukhin, I. (2017).
\newblock Attention is all you need.
\newblock In {\em Advances in neural information processing systems}, pages
  5998--6008, Long Beach, USA.

\bibitem[Wu and Wang, 2007]{wu2007pivot}
Wu, H. and Wang, H. (2007).
\newblock Pivot language approach for phrase-based statistical machine
  translation.
\newblock {\em Machine Translation}, 21(3):165--181.

\bibitem[Zhang et~al., 2019a]{zhang-etal-2019-aspect}
Zhang, C., Li, Q., and Song, D. (2019a).
\newblock Aspect-based sentiment classification with aspect-specific graph
  convolutional networks.
\newblock In {\em Proceedings of the 2019 Conference on Empirical Methods in
  Natural Language Processing and the 9th International Joint Conference on
  Natural Language Processing (EMNLP-IJCNLP)}, pages 4568--4578, Hong Kong,
  China.

\bibitem[Zhang et~al., 2019b]{Zhang2019aSentClass}
Zhang, C., Li, Q., and Song, D. (2019b).
\newblock Syntax-aware aspect-level sentiment classification with
  proximity-weighted convolution network.
\newblock In {\em Proceedings of the 42nd International ACM SIGIR Conference on
  Research and Development in Information Retrieval}, SIGIR’19, page
  1145–1148, Paris, France.

\end{thebibliography}

\end{document}